\documentclass{INTERSPEECH2023}

\usepackage{tabularx}

\interspeechcameraready

\title{The IMS Toucan System for the Blizzard Challenge 2023}
\name{Florian Lux, Julia Koch, Sarina Meyer, Thomas Bott, Nadja Schauffler,\\Pavel Denisov, Antje Schweitzer, Ngoc Thang Vu} 
\address{University of Stuttgart, Germany}
\email{florian.lux@ims.uni-stuttgart.de}

\begin{document}

\maketitle
 
\begin{abstract}
For our contribution to the Blizzard Challenge 2023, we improved on the system we submitted to the Blizzard Challenge 2021. Our approach entails a rule-based text-to-phoneme processing system that includes rule-based disambiguation of homographs in the French language. It then transforms the phonemes to spectrograms as intermediate representations using a fast and efficient non-autoregressive synthesis architecture based on Conformer and Glow. A GAN based neural vocoder that combines recent state-of-the-art approaches converts the spectrogram to the final wave. We carefully designed the data processing, training, and inference procedures for the challenge data. Our system identifier is G. Open source code and demo are available.
\end{abstract}


\section{Introduction}
The 2023 installment of the Blizzard Challenge is concerned with the generation of natural speech in the French language given two datasets of female French speakers. The setup is split into two tasks, the hub task and the spoke task. The goal in the hub task is to create a French speech synthesis system of as high quality as possible, using only publicly available resources and models. 
In the spoke task, there is no such restriction, but the goal is to produce speech that is as similar as possible to the speaker of the corresponding data set without losing the greatest possible naturalness.
This is especially challenging, since the amount of data available for the spoke task is small compared to standard datasets used for the text-to-speech (TTS) task. The datasets for the hub task and for the spoke task are both single speaker French datasets by different female native French speakers from France. The dataset for the hub task consists of five audiobooks from LibriVox\footnote{\url{https://librivox.org/}}, with a total of 289 chapters and 51 hours of read speech by Nadine Eckert-Boulet (NEB). The dataset for the spoke task is also read speech but much shorter (2 hours of speech) and read by the speaker Aurélie Derbier (AD). 60\% of the utterances are read from different books while the rest comes from parliament transcripts. Besides the audios, the datasets contain text and phonetic alignments to 2/3 of the utterances. 
After the submission of the TTS systems for each task, subjective listening tests were conducted by the challenge organizers to assess the performance of each system. For both tasks, the quality and speaker similarity were evaluated on speech synthesized from book sentences as mean opinion scores (MOS). For the hub task, an intelligibility test was conducted, for which raters had to transcribe an utterance. Further, the correctness of homograph pronunciations was measured. 
The listening tests were performed by both paid native speakers and volunteers, with a distinction between speech experts and naive listeners. 

We improve on our previous submission to the Blizzard Challenge 2021 \cite{lux2021toucan}. The largest portion of the changes we made is the culmination of the last 2 years of work on the IMS Toucan toolkit, which introduces a lot of designs to handle multilinguality, controllability and low-resource scenarios. We call the system where all those designs come together ToucanTTS and enrich it with additional components that are specific to this challenge. Our system identifier is G. The exact code used, as well as an interactive demo, is available open source\footnote{\url{https://github.com/DigitalPhonetics/IMS-Toucan}}. 

\section{Architecture}
\begin{figure*}
    \centering
    \includegraphics[width=.9\textwidth]{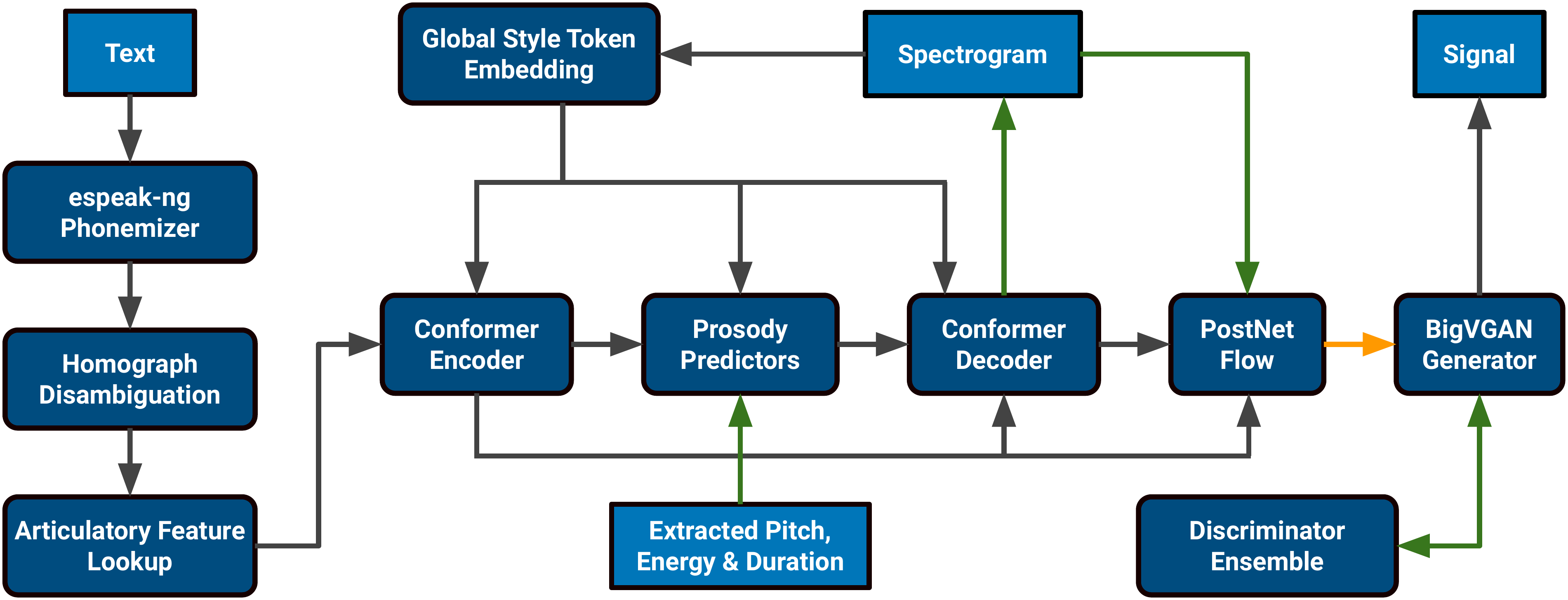}
    \caption{Overview of all the components in our system. The green arrows show the losses applied at training time. The orange arrow only exists during inference, the gradient is not passed through at training time.}
    \label{fig:overview}
\end{figure*}

An overview of our system is shown in Figure \ref{fig:overview}. Data-efficiency is one of the key objectives of our IMS Toucan toolkit. We split the generation process into many small steps so that the individual subtasks are easy for a model to learn. This means that comparably few parameters are required, which greatly reduces the need for training data. 
This is certainly the main advantage of our toolkit. However, data-efficiency comes at the expense of synthesis quality and naturalness, which is confirmed by the evaluation. 
Our goal in this challenge is to see, how well our data-efficient, fast, lightweight, and highly controllable approach can perform compared to systems which are purely optimized for quality and naturalness.

\subsection{Text-to-Phoneme} \label{sec:textfrontend}
To convert any text into a sequence of phonemes, we use an open source phonemizer\footnote{\url{https://github.com/bootphon/phonemizer}} with espeak-ng\footnote{\url{https://github.com/espeak-ng/espeak-ng}} as its backend. We perform rudimentary text cleaning and then transform the input into a sequence of phonemes using the IPA notation. These phonemes are then replaced by articulatory vectors, like we introduced in \cite{lux2022language}. This means, that each phoneme is transformed into a vector, that contains a one-hot-encoding of the configuration of the human vocal tract while producing this sound through a lookup table. These representations include additional dimensions to account for additional nonsegmental markers, like lenthening, shortening, and lexical stress. The phonemizer also produces those symbols, which do not make up units on their own, but instead they modify preceeding or following units, changing the value of the respective dimension in the articulatory vector. This extension was made to our system in \cite{lux2022low}, in order to be able to account for tonal languages, in which tone is another crucial instance of nonsegmental markers in the phoneme sequence.

\subsection{Spectrogram-to-Alignment} \label{sec:aligner}
Our approach relies on precise alignments of the phonemes to the spectrogram frames, not only because of the durations that the model learns themselves, but also because the pitch and energy values we are using are averaged over phoneme durations. To get those precise alignments, we train a simple speech recognition system using a CTC objective \cite{graves2006connectionist}, that models the likelyhood of all phonemes over time. This posteriogram is then fed into an auxiliary spectrogram reconstruction model, that tries to reconstruct the inputs in order to make sure that the borders between the phonemes are more sharply defined. We introduced this component to our toolkit in \cite{lux2023exact} and verified its accuracy and usefulness in \cite{PoeticTTS, meyer2023prosody}. Another recent paper that verifies the accuracy of such a system is \cite{zalkow2023evaluating}. To get the alignments from the posteriograms, we re-order the axis containing the phoneme likelyhoods by the order of phonemes in the transcription and run a monotonic alignment search (MAS). We compare this to a path-search using the Dijkstra algorithm, however we find that the Dijkstra algorithm tends to skip over some segments, which the MAS is constrained not to do. 

\subsection{Spectrogram-to-Embedding}
To disentangle and capture varying acoustic conditions as well as speaking styles, 
despite there being only two single speakers in the challenge data, we use the Global Style Token embedding approach \cite{wang2018style} together with some augmentations recommended in AdaSpeech 4 \cite{Wu2022AdaSpeech4A}, namely a style token disentanglement loss and increasing the amount of style tokens to 2000. The embeddings are integrated after every encoder block, decoder block and every layer in the prosody predictors using concatenation followed by projection.

\subsection{Phoneme-to-Spectrogram}
As our spectrogram generation network, we employ the basic structure of FastSpeech 2 \cite{ren2020fastspeech}, augmented with phoneme-wise averaging of pitch and energy following FastPitch \cite{lancucki2021fastpitch}. This combination allows us to achieve a remarkable level of fine-grained control over the generated speech. For efficiency, we chose the Conformer architecture \cite{gulati2020conformer} as encoder and decoder, which excels at many speech tasks. To further enhance the system, we incorporate a PostNet using normalizing flows, inspired by the analysis presented in PortaSpeech \cite{ren2021portaspeech, kim2020glow}. The model consists of roughly 46,000,000 trainable parameters. The implementation on which we base our challenge system is the release associated with \cite{lux2023controllable} in our open-source toolkit \cite{lux2021toucan}.

\subsection{Spectrogram-to-Wave}
As the neural vocoder to perform spectrogram inversion, we use a generative adversarial network (GAN) \cite{gangoodfellow} setup consisting of the BigVGAN generator \cite{lee2022bigvgan} together with the discriminators introduced in MelGAN \cite{kumar2019melgan}, HiFiGAN \cite{kong2020hifigan}, and Avocodo \cite{bak2022avocodo}. Following the analysis of the winner of the 2021 Blizzard Challenge \cite{liu2021delightfultts}, we use 16kHz spectrograms to make the task easier for the spectrogram generator, but then 
perform superresolution together with spectrogram inversion by using upsample scales that map to a 24kHz waveform.

\section{Data Preprocessing}
\subsection{Data Splitting}
We split the chapter-wise audios of the challenge data at the paragraph boundaries using the provided alignments. To ensure that a sentence break within an utterance would not be unseen and therefore not cause bad prosody at inference time, we made a set of even longer utterances by concatenating consecutive sentences into joint utterances with a short pause of 0.22 seconds inbetween. Longer form audios where this design might be helpful were included in the submitted test data, however not part of the evaluation of the challenge. Each utterance was concatenated with the subsequent utterances so that the duration of the joint utterance does not exceed 15 seconds including the inserted pauses. 
The number of the generated joint utterances was 1640 for AD and 7967 for NEB, the number of subsequent sentences in these utterances was up to 5 for AD and up to 6 for NEB.

\subsection{Signal Processing}
We perform loudness normalization on all data 
using the pyloudnorm tool \cite{steinmetz2021pyloudnorm}. We normalize the loudness of the training data to -30dB and then adapt the loudness of our system output during inference to precisely match the loudness of the human references. Since the quality of the challenge data was not on a clean studio level and included artifacts, background sounds, and a lot of reverb, we chose to perform speech enhancement on the AD dataset. For this we used the Adobe Podcast Enhance software through its web interface\footnote{\url{https://podcast.adobe.com/enhance}, accessed April 2023}, which is free to use, however neither open-source nor currently accessible trough an off-the-shelf API. Thus we did not apply the enhancement step to the NEB data, as per the challenge rules of the Hub Task. We explored open-source alternatives to perform speech enhancement on the NEB data as well, however we found none with satisfactory results and therefore used the NEB data as it was given without further processing. 

\subsection{Feature Representation}
As intermediate feature representation in our system we chose a log-mel-scaled spectrogram. Although several approaches using end-to-end architectures \cite{kim2021conditional, donahue2020end} as well as architectures that rely on neural audio codec representations \cite{chen2023vector, shen2023naturalspeech} have shown that the spectrogram is not an ideal representation for use in a speech synthesis setting,  we still currently prefer spectrograms due to their interpretability and reliability. In the future however, with more and more neural audio codecs being developed rapidly at this point in time, we are planning to exchange the spectrograms for a different representation. The settings we use to extract spectrograms are as follows: We calculate a spectrogram on a 16kHz waveform with a window size of 1024 and a hop length of 256 and a Hann window. We then transform the spectrogram into a mel-spectrogram with 80 frequency bins. Finally we apply a log with base 10 to make the value ranges easier to reconstruct for the synthesis model. We use the Librosa toolkit \cite{mcfee2015librosa} to extract the features.

\subsection{Prosody Representation}
Since our system models the prosody of an utterance explicitly using 
phone durations, pitch values per phone and energy values per phone, we 
explored reliable ways of extracting these values with great accuracy. To extract durations, we use the aligner described in section \ref{sec:aligner}. Pitch and energy are extracted with praat-parselmouth \cite{parselmouth}. We average pitch and energy values for each phone as proposed in \cite{lancucki2021fastpitch} to achieve controllability at phone-level. Further, we manually set pitch values of all unvoiced phones as well as pitch and energy values of silent symbols such as pause markers and punctuation symbols to zero to reduce noise. The pitch and energy levels are normalized by the mean per utterance excluding zeros to make prosody curves speaker independent.

\subsection{Pretraining Datasets} \label{subsec:pretraining}
As both speakers of the challenge are female, we chose to use a subset of only female speakers in all pretraining data. Further, since both speakers speak French as it is spoken in France, we tried to heuristically exclude speakers from other regions where French is spoken, such as Canada, by removing speakers from the Multilingual LibriSpeech (MLS) corpus \cite{Pratap2020MLSAL} who also recorded audio books in languages other than French on LibriVox. We also excluded the sessions of the speaker NEB who is included in MLS as well. The datasets we used are shown in Table \ref{tab:datasets}. The SIWIS French Speech Synthesis Database \cite{siwis} consists of high quality French audio, spoken by one female French speaker. From this corpus, we exclude the chapter reading (part 4 and 5) because they have not been published in a segmented form. We further considered including VoxPopuli \cite{wang-etal-2021-voxpopuli} and 
the French as spoken in France subset of the Phonologie du Fran\c{c}ais Contemporain (PFC) database \cite{11403/pfc/v1} which is a research collection of various French accents across the globe. Eventually we decided against using VoxPopuli and PFC, since both contain mixed quality recordings, which reduced the overall quality of speech produced by our system. Generally, our system tends to perform best given the cleanest data, even if not much is available. 
We also investigated the SynPaFlex dataset \cite{sini-etal-2018-synpaflex} but had to remove it because it almost exclusively consists of utterances by the speaker NEB, which would violate the challenge rules. 

\begin{table}[]
    \centering
    \begin{tabular}{lll}
        \toprule
        Dataset & \# Speakers & Hours \\
        \midrule
        Multilingual LibriSpeech \cite{Pratap2020MLSAL} & 56 & 157 h 51 min\\
        SIWIS \cite{siwis} & 1 & 10 h 2 min\\
        \midrule
        NEB & 1 & 51 h 12 min\\
        AD & 1 & 2 h 3 min\\
        \bottomrule
    \end{tabular}
    \caption{Datasets used to train the system, with the subsets as described in Section \ref{subsec:pretraining}. The top rows are used for pretraining and the bottom rows for finetuning for the respective challenge tasks.}
    \label{tab:datasets}
\end{table}

\subsection{Homograph Resolution}
We implement a rule-based disambiguation step on top of espeak to detect and resolve homographs. For this purpose, we extracted a list of 800 French homographs from Wiktionary\footnote{\url{https://fr.wiktionary.org/wiki/Cat\%C3\%A9gorie:Homographes_non_homophones_en_fran\%C3\%A7ais}} and stored them in a dictionary together with their phonetic transcriptions and POS tags. This is already sufficient to disambiguate a huge amount of homographs, where different pronunciations also correspond to different POS tags (e.g. \textit{adoptions}: \textipa{\textbackslash ad\textopeno psj\~\textopeno\textbackslash}[NOUN] - \textipa{\textbackslash ad\textopeno ptj\~\textopeno\textbackslash}[VERB]). However, there remain a number of homographs where different pronunciations have the same POS tag. In these cases, we enrich our annotations with more fine-grained POS tags that also contain morphological information such as gender, number or tense \cite{labrak:hal-03696042}. This allows us for example to distinguish the singular form of the word \textit{fils} from the still ambiguous plural forms, i.e. wherever \textit{fils} is tagged as singular, we know the correct pronunciation is \textipa{\textbackslash fis\textbackslash}. We further notice that this procedure often distinguishes anglicisms from French native words, which are however not evaluated in the challenge. 
If there is still ambiguity left, we decide for one default pronunciation based on which word sense we expect to be more likely. 

During inference, we tokenize and POS tag the input text with the POET tagger \cite{labrak:hal-03696042} that uses the same extended tagset as mentioned above. We take the configuration of the POET tagger using Flair \cite{akbik2018flair} and CamemBERT \cite{martin2020camembert} embeddings as input for a Bi-LSTM with Conditional Random Field (CRF) for sequence tagging\footnote{\url{https://huggingface.co/qanastek/pos-french-camembert-flair}}. For each token, we then check whether it occurs in our dictionary of homographs. If so, we look up the correct pronunciation for the current homograph given its POS tag. Since we did not annotate all entries in our homograph list with the extended POS tag, we first map the tags given by the POET tagger back to the coarse POS tags extracted from Wiktionary and check if there is already an unambiguous entry for the current instance. If not, we check if there is an entry in the dictionary with a matching annotation of an extended POS tag. If there is no matching candidate, we fall back to the default pronunciation.

In addition, we decided to handle the very frequent word \textit{plus} separately with regular expressions according to French grammar rules. We first check if \textit{plus} occurs in the context of a negation and thus, has a negative meaning in the sense of \textit{no more}. In this case, it is always pronounced \textipa{\textbackslash ply\textbackslash}. Else if \textit{plus} is used in a positive meaning (i.e. meaning \textit{more}), usually the pronunciation is \textipa{\textbackslash plys\textbackslash}. However, only if none of the following exceptions apply: 1. If the following word is an adjective or adverb that starts with a consonant, the correct pronunciation is \textipa{\textbackslash ply\textbackslash}. 2. If the following word is an adjective or adverb and starts with a vowel, we have to take care of the \textit{liaison}, and thus, the pronunciation is \textipa{\textbackslash plyz\textbackslash}.


\subsection{Silence Annotation}
As heuristic markers for pauses, we use the characters [, ; - "]. Since speaker AD made pauses very inconsistently with those markers, we use the durations from the aligner and an open-source tool for voice activity detection \cite{SileroVAD}, to check if each occurrence of a pause marker actually corresponds to a silent segment in the signal and remove the pause marker from the transcript otherwise.

\subsection{Data Cleaning}
We calculate the loss for each individual sample in the datasets and remove the samples with the highest loss until the average loss of the next 10 samples in the ranking is no more than 0.1 smaller than the highest. This removes a few mispronunciations from the training data, as well as samples with coughing noises or laughter in the background for AD.

\section{Training Procedure}
To increase robustness, we decided to include a pretraining stage on large and diverse French datasets described in section \ref{subsec:pretraining}, followed by a finetuning stage on each of the challenge datasets. We used the same hyperparameters for both the pretraining and the finetuning, except for the amount of steps trained. Typically, learning rates are reduced during finetuning, which we chose not to do, since we only care about the performance on the finetuning data. As optimizer we used Adam \cite{kingma2014adam} with default settings except for the learning rate of 0.001. We included 8,000 steps of warmup, used a batchsize of 24, and started updating the postnet flow after 20,000 steps. Pretraining ran for 80,000 steps, finetuning for NEB was stopped after 40,000 steps, finetuning for AD was stopped after 30,000 steps. We trained the GST parameters during pretraining, but then froze those parameters during finetuning. For validation, we held out 10 samples summing up to one minute from each speaker. The vocoder was trained from scratch for 1,000,000 steps on all the French data we accumulated. 

\section{Inference Procedure}
\subsection{Inference Speed}
During inference, our system is able to synthesize 24 seconds of audio per second on an NVIDIA RTX A6000 GPU without using batching and 2 seconds of audio per second on an Intel i7 9700k CPU without using batching. 

\subsection{Style Reference}
Since our System is trained with GST, we have to supply a reference audio at inference time. Since the test sentences cannot influence our choice, we carefully selected for each speaker multiple candidate utterances for speaker embeddings that exhibit distinct speaking styles and microphone and room characteristics. In an internal A/B testing round, we decided on the embedding for each speaker that was perceived as the most pleasant to listen to by majority voting to be used in all cases.

\subsection{Signal Processing}
We generate audio at 24kHz and then double every value on the time axis to get to a more standard 48kHz. To avoid imaging issues, we apply a low-pass filter at 12kHz with a roll-off of 6dB per octave. We adjust the loudness to -29dB for NEB and -33dB for AD using pyloudnorm. The encoding used is int16 for maximized compatibility.

\section{Unsuccessful Designs}
\subsection{Adversarial Feedback}
To investigate whether the naturalness gap to end-to-end models is in part caused by the feedback signal being adversarial, we replaced the L1 distance with a discriminator that is trained to distinguish real and and generated spectrograms alongside the TTS. In internal A/B testing, we did not notice a significant improvement and therefore left this design out of our final submission.

\subsection{Word and Sentence Embeddings}
We experimented with word and sentence embeddings as additional conditioning signals. Sentence embeddings capture semantic and structural properties of a sentence and contextual word embeddings contain information about the meaning and importance of each word. Therefore they can be helpful for TTS systems to produce more natural prosody and pronunciation \cite{Hayashi2019}. 
We used a pretrained sentence transformer model \cite{reimers2019sentence} based on CamemBERT \cite{martin2020camembert} \footnote{\url{https://huggingface.co/dangvantuan/sentence-camembert-base}} to extract sentence embeddings from input text. In our model architecture they were first passed through adaptation layers and then concatenated with the GST embeddings. In internal A/B testing we found no significant improvements when using sentence embeddings.
Word embeddings were extracted from a pretrained CamemBERT model \footnote{\url{https://huggingface.co/camembert-base}} by combining the last 4 hidden layers for each sub-token and then averaging embeddings of sub-tokens that form words. In our TTS system we concatenated each word embedding with its corresponding phoneme embeddings according to their word boundaries.
In earlier training stages the word embeddings helped the model not to make mispronunciations. However, after a longer training time, we found no significant differences in internal A/B testing and therefore also left this design out of our final submission.

\subsection{Variational Variance Predictors}
The deterministic variance predictors are a major bottleneck for the naturalness and livelyness of the prosody of generated speech. To alleviate this, we explored the use of various generative methods to generate realistic and variable pitch, duration and energy curves for a given sentence and style embedding. We explored the use of a variational auto encoder (VAE), which is however difficult to implement since the 1D nature of the data does not allow for an information bottleneck, without which VAEs do not work. We built a GAN for producing these curves instead, but it would not converge, even if a Wasserstein distance \cite{wgan} was employed as cost function. Finally we built variance predictors based on normalizing flows. These performed well and lead to interesting yet natural sounding prosody. However, those components were not very stable and required frequent resets to prior checkpoints during training whenever they collapsed and their loss values exploded. We could not find a solution to stabilize training in time for the submission, but still believe that this approach is worth pursuing further.

\subsection{Speech Enhancement}\label{subsec:enhancement}
In our final system, we used a speech enhancement model on the speech of speaker AD and left the speech of speaker NEB untouched due to the challenge rules. We tried to find a speech enhancement model, that would fit within the challenge rules with which we could remove the reverb from the NEB dataset and clean up the varying microphone qualities used across multiple datasets. For this, we used VoiceFixer \cite{liu2021voicefixer}, however the result sounded not convincing enough to go forward with it. An open-sourced speech restoration model with performance close to the proprietary models like e.g. Miipher \cite{koizumi2023miipher} might have boosted the sound quality of our system significantly.

\section{Challenge Results}
Our performance for selected categories in the hub task is shown in Table \ref{fig:hub}. Our performance in the spoke task is shown in Table \ref{fig:spoke}. Overall, our submission was ranked among the lowest scoring systems across most tasks. So our fast, efficient and controllable approach seems to lack in the general synthesis naturalness and quality. With this insight gained, we aim to address this bottleneck of our approach in the future.

\begin{table}[h]
    \centering
    \begin{tabularx}{\linewidth}{l|X|X|X}
    \toprule
        Task & Better & Equal & Worse  \\
        \midrule
        \textbf{Hom} & & J, M, Q, H, T, F, O & D, N, S, BF, C, L, P, K, E, R, BT, I  \\\midrule
        \textbf{Nat} & A, F, I, O, M, P, Q, T, J, E, S, H, D, C, K, L & R, N & BF, BT  \\
        \bottomrule
    \end{tabularx}
    \caption{\textbf{Hom}ograph correctness and \textbf{Nat}uralness of other systems relative to ours in the \textbf{hub} task according to significance tests. A is human speech.}
    \label{fig:hub}
\end{table}

For the homograph disambiguation task, we rank top 5 with an accuracy of 84\%, while no other system is significantly better than ours. This shows that a simple rule-based system built on linguistic expert knowledge can handle the task sufficiently well. Considering the homograph \textit{plus}, we notice that our system sometimes produces incorrect pronunciation although the phoneme input to the TTS model is correct. We attribute this to the fact that we apply homograph disambiguation at inference time only, so the model learned to ignore the fine difference between \textipa{\textbackslash ply\textbackslash}, \textipa{\textbackslash plys\textbackslash} and \textipa{\textbackslash plyz\textbackslash} due to incorrect phoneme input during training. 

\begin{table}[h]
    \centering
    \begin{tabularx}{\linewidth}{l|X|X|X}
    \toprule
        Task & Better & Equal & Worse  \\
        \midrule
        \textbf{Nat}& F, A, O, L, Q, H, J, P, T, E, S & BF, R & N, K, BT  \\\midrule  
        \textbf{Sim}& Q, F, J, L, P, A, E, H, T, S, O, BF, R, N, BT, K & &  \\
        \bottomrule
    \end{tabularx}
    \caption{\textbf{Nat}uralness and \textbf{S}imilarity of other systems relative to ours in the \textbf{spoke} task according to significance tests. A is human speech.}
    \label{fig:spoke}
\end{table}

Our bad scores in the similarity evaluation can most likely be attested to the use of the speech enhancement used. The enhancement model altered the voice slightly, but still improved the naturalness greatly, since the TTS had problems with the reverb otherwise. Interestingly, our speaker similarity scores look much better when looking at the subset of non-native French speakers. This aligns with our perception, since no one from our team speaks French and we decided in favor of the enhancement in internal A/B testing.

\section{Conclusion}
Our system is able to handle unseen phonemes due to the articulatory features used, it is highly controllable due to the FastPitch style averaging of pitch and energy used, it is fast due to the use of the fully parallel conformer as encoder and decoder and it is robust due to the non-autoregressive nature of the decoding. The overall naturalness of the prosody however is bottlenecked due to the deterministic pitch, energy and duration predictors. Further, the two-step synthesis procedure using spectrograms as intermediate representations is known to be very data efficient, however not as natural as systems trained fully end-to-end. We plan to address both of these shortcomings in a future version of our system using stochastic predictors for the variance in the speech signal and neural audio codecs to replace the spectrograms as intermediate representations to retain the data efficiency of our approach. Overall, we believe that despite the low ratings in naturalness that we achieved in the challenge, the benefits that our system offers w.r.t. speed, data efficiency, robustness and controllability are still very valuable assets to have in a TTS system, and we need to put in further work to enhance the naturalness of the generated speech without compromising the other desireable properties of our system. For the French language specifically, our carefully designed rule-based homograph disambiguation together with the rule-based espeak phonemizer performs the text-to-phoneme conversion very well, despite the simplicity of a rule based system, indicating that this task can still be handled very well by simple systems and does not necessarily need large neural models, if sufficient linguistic knowledge for the language is available.

\bibliographystyle{IEEEtran}
\bibliography{main}

\end{document}